\documentclass[journal]{IEEEtran}
\usepackage{amsmath,amssymb}
\usepackage{mathtools}
\usepackage{multirow}
\usepackage[export]{adjustbox}

\usepackage{graphicx}
\usepackage{dcolumn}
\usepackage{bm}
\usepackage{hyperref}
\usepackage[english]{babel}
\usepackage[normalem]{ulem}
\usepackage{comment}
\usepackage[draft]{todonotes}   
\usepackage{xcolor}
\usepackage{booktabs}

\DeclareMathOperator*{\argmax}{arg\,max}

\title{Learn Like The Pro: Norms from Theory to Size Neural Computation\thanks{Corresponding Author: ravela@mit.edu}}%

\author{\IEEEauthorblockN{Margaret Trautner}\\\IEEEauthorblockA{\it Department of Computation and Mathematical Sciences\\ California Institute of Technology, Pasadena, CA}\\
\IEEEauthorblockN{Ziwei Li}\\
\IEEEauthorblockA{\it Earth, Atmospheric and Planetary Sciences\\ Massachusetts Institute of Technology, Cambridge, MA}\\
\IEEEauthorblockN{Sai Ravela}\\\IEEEauthorblockA{\it Earth Signals and Systems Group\\ Earth, Atmospheric and Planetary Sciences\\ Massachusetts Institute of Technology, Cambridge, MA}}
\begin{document}
\maketitle

\begin{abstract}
The optimal design of neural networks is a critical problem in many applications. Here, we investigate how dynamical systems with polynomial nonlinearities can inform the design of neural systems that seek to emulate them. We propose a Learnability metric and its associated features to quantify the near-equilibrium behavior of learning dynamics. Equating the Learnability of neural systems with equivalent parameter estimation metric of the reference system establishes bounds on network structure. In this way, norms from theory provide a good first guess for neural structure, which may then further adapt with data. The proposed approach neither requires training nor training data. It reveals exact sizing for a class of neural networks with multiplicative nodes that mimic continuous- or discrete-time polynomial dynamics. It also provides relatively tight lower size bounds for classical feed-forward networks that is consistent with simulated assessments. 
\end{abstract}

\section{Introduction}

Numerical simulation of nonlinear dynamical systems is central to many applications. However, high-fidelity simulations often demand highly resolved dynamics and variables, which can be computationally prohibitive. Thus,  it is common practice to filter/truncate governing equations, decompose the computational domain, or coarsen space/time resolution to control efficacy, all of which limit predictive skill. 

The apparent ability to capture nonlinear relationships and rapidly execute has brought significant attention to Neural Networks. For dynamical systems, they serve as stand-alone or hybrid models, parameterizations of unrepresented features, or as surrogates or emulators of numerical models of governing equations.  Interest has surged, for example, in using neural networks to emulate chaotic systems {\cite{Bakker2000, Dudul2005, Bahi2012, Zerroug2013, Zhang2017, Yu2017, Madondo2018}}. Many nonlinear dynamical systems are chaotic~\cite{Strogatz2015}; minor variations in initial conditions produce divergent solutions~\cite{Lorenz1963}.  It turns out that Neural Networks are pretty successful at mimicking chaos~\cite{Li2019bc}. They learn topological mixing, a characteristic of chaos, and demonstrate near-identical predictability to the chaotic system they train to emulate.  

These successes make it easy to forget that we understand very little about Neural Networks. Creative designs for deep neural networks often lack a firm scientific basis. One fundamental problem is  {\em sizing}: how many hidden nodes and layers might the network need? 

Variable elimination is one approach to tackle this problem; start with an extensive network and promote sparsity to prune the weights~\cite{hoefler21}. Another approach is to sweep the size parameter and use cross-validation or other model selection procedures to grow the network~\cite{hoefler21,raschka18}. Time permitting, one could even resort to Markov Chain Monte Carlo~\cite{lee2012}. These approaches are often essential in practice, but they entail unsustainably excess computation. Therefore, a pretty good size first guess prior to entertaining additional optimization is highly desirable. 

Tight bounds on network size are difficult to obtain. One approach uses the dataset to estimate network size either directly or through an easier-to-train machine to gauge the ``nonlinearity" and ``dimensionality" in the data~\cite{lappas07,stathkis09,friedland2018practical,dsouza2020}. In contrast, using the governing equations for sizing when extant theory is available is particularly desirable. Because theory is often approximate and governing equations are generally imperfect (e.g., due to numerics), networks sized appropriately can further adapt quickly, such as by using informative structure learning~\cite{trautner2020}. 

We are interested in sizing neural dynamical systems~\cite{trautner2019,trautner2020} using the available governing equations. In particular, this paper focuses on single-layer neural networks that emulate the cruical and large class of autonomous dynamical systems with polynomial nonlinearities. The Universal Approximation theorem~{\cite{Hornik1989,  Hornik1991, Seidl1991, Funahashi1993}} further broadens the scope of single-layer studies. The present article further restricts the governing dynamics to ordinary differential equations and trains using gradient descent procedures with least-squares loss. 

Within this scope, the central claim of this paper is that known governing dynamics lower bound network size without requiring data, training, or an explicit error budget. To establish the claim, we show that the near-equilibrium behavior of a reference system constrains the size of the neural network that seeks to match the reference system's behavior. Near equilibrium is defined in the sense of stability under small perturbations of system parameters. The intuition behind the approach is that the reference system is a pretty good if not ideal design. The agility with which it recovers from minor parameter variations is an essential feature for a learning system to match. And, if the learning system is just as agile, its admissible sizes are unlikely to be very varied. This paper shows that this idea works surprisingly well. It matches independent experimental results on size bounds and provides  tighter lower size bounds than extant work that requires data size, error tolerances, and/or the detailed system design as inputs. 

Our approach first calculates a suitable norm from the reference system's parameter gradients (the parameter Jacobian) using bounded system inputs and the known polynomial equations. It then estimates the corresponding near-equilibrium norm for the learning system. Because the norms are explicit functions of size variables, matching them constrains the unknown learning system size. We show that for the class of PolyNets~\cite{trautner2019}, exact networks for polynomials, network sizes are trivially determined. Thus linear network sizes are also precisely determined with bounded inputs and lower bounds nonlinear network sizes. A Bayesian approach further tightens the lower size bounds for nonlinear networks. Random network simulations calibrate the nonlinear network norms with corresponding linear networks and, in this way, derive tighter size lower-bounds from the reference system. Our sizing results extend to both continuous-time polynomial ordinary differential equations and numerical discrete-time difference equations. 

We demonstrate applicability using the Lorenz-63 (L63) model {\cite{Lorenz1963}}, which is a low-dimensional dissipative chaotic dynamical system. The lower bounds derived suggest that the multiplicative PolyNet requires two hidden nodes. A linearly activated additive network armed only with the system dimension $n=3$ and polynomial degree $d=2$ requires four nodes, a residual network requires three nodes, and a nonlinearly activated network requires six nodes. These results are remarkably consistent with experimentally derived bounds by explicit training described elsewhere~\cite{Li2019bc}. 

The remainder of this paper is organized as follows. In Section~\ref{sec:rw}, related work is presented. In Section~\ref{sec:elt} presents the Learnability metric, which provides for a sizing method in Section~\ref{sec:msf}. The bounds further improve for general one-layer networks in Section~\ref{sec:rns}. This paper's conclusions follow in Section~\ref{sec:concl}.

\section{Related Work}
\label{sec:rw}

The literature describes several approaches to size networks. Stathkis~\cite{stathkis09} presents a near-optimal solution to sizing in the context of classification in remote sensing, using a genetic algorithm. The fitness function concurrently seeks the most accurate and compact
solution. In recent literature, based on MacKay's information-theoretic model of supervised machine learning~\cite{mckayinfth}, Friedland et al.~\cite{friedland2018practical} estimate the maximum size of a neural network given a training data set that includes the analytic estimation of network capacity and a heuristic method to estimate the neural network capacity required for a given dataset and labeling.  Lappas et al.~\cite{lappas07} argue, as we have here, that many theoretically derived bounds are unrealistic. They develop a sizing procedure using only the available training data size, achieving a high classification rate. D'Souza et al.~\cite{dsouza2020} point out that in the "Small Data" challenge, the "optimal network" is data-driven and not just based on size, which they show using the VC-dimension's~\cite {blumer89} influence on structural hyperparameters. Optimizing over "all possible combinations" of structure hyperparameters, they show marked CNN improvements on MNIST classification. Reliance on Valiant's notion of Learnability~\cite{Valiant1984} in conjunction with the VC dimension has also led to estimates of neural capacity~\cite{blumer89,LINIAL199133}. 

In contrast to the methods mentioned above, this paper takes an alternate route to Learnability. In contrast to VC-dimension or Vapnik's Learnability, stability under parameter estimation (or learning) dynamics is the key criterion. Stability is closely related to Lyapunov exponents~\cite{Strogatz2015} of the Learning system in training and associated measures such as Lyapunov sums. Local metrics of Learnability establish, in our case, the agility with which specific directions of error are reducible (to first order), and integrated versions of these local metrics lead to connections with Finite-Time Lyapunov Exponent~\cite{Haller2001DistinguishedFlows,brunton2010,SHADDEN2005271} as Learnability features. We claim that this provides tighter bounds by better quantifying and exploiting near-equilibrium learning behavior.  Related to this is that we have a reference model, which may not always be available. Thus, data is not needed, and neither is any explicit training -- the near-equilibrium behavior of the reference model concerning its parameters serves as a guide to sizing. 

Polynomials have been modeled as neural networks {\cite{Andoni2014}}, therefore, we examine some bounds based on prior results. In particular, if we model a polynomial as a two-layer network with one output, $y = \sum_{j} a_j \phi(\mathbf{w}_j \mathbf{x})$, with a linear part $\mathbf{w}_j \mathbf{x}$ on input $\mathbf{x}$, and a smooth activation function $\phi$, then certain results are immediately accessible. 

For dense polynomials trained by gradient descent, it has been shown that a network for degree $d\geq 1$ and input $n$ with convergence to $\epsilon>0$ using $m=\Omega(n^{6d}/\epsilon^3)$ hidden nodes and learning rate $\lambda < \frac{1}{4m}$ needs $O(\frac{n^{2d}}{\lambda\epsilon^2 m})$ time-steps and $m^{O(1)}$ samples {\cite{Andoni2014}}. This is rather poor compared to regression and could, in principle, argue that estimating the parameters of a quadratic polynomial for each rate term would suffice. However, when the system is \textit{k-sparse}, other results {\cite{Barron1993}} suggest that convergence with $O(nk\cdot d^{O(d)})$ hidden nodes is feasible. As we shall see, even these results are weaker than the bounds derived here, also see~\cite{Li2019bc}.

\section{The Learnability Metric}
\label{sec:elt}
Equilibrium Learning Dynamics refers to the dynamics of the learning system as it trains near the vicinity of parameter convergence, local or global. We are interested in comparing systems based on their variational behavior around an equilibrium. A {\em Learnability}\footnote{Learnability is used in the sense of spectral properties of metrics induced by the error dynamics during learning. It is different from, e.g., Valiant's definition\;{\cite{Valiant1984}}.} metric enables this, and the next section shows its use for network sizing. In this section, we develop the necessary foundations. 

Consider a trained neural network of the form: 
 \begin{eqnarray}
     y_s &=& \hat{y}^*_s +e^*_{s} \\\nonumber
         &=&f(x_s;w^*)+e^*_{s} 
 \end{eqnarray}
 where, $w^*$ are the optimal weights (network parameters), $\hat{y}^*_{s}$ is the prediction with $w^*$ and sample $s$ input vector  $x_s$. The vector $e^*_{s}\sim \mathcal{N}(0,R)$ is the intrinsic prediction error with respect to truth $y_s$, and it is assumed {\it wlog} to be zero-mean Gaussian with  covariance $R$. Let's call this the learning equilibrium.
 
 Consider a infinitesimal perturbation $w+\delta w = w^{*}$ away from the learning equilibrium. Assuming the Taylor expansion exists to first order, we  obtain:
\begin{eqnarray}
   e_s &=& f(x_s,w^*)-f(x_s,w)\\
   &=& \nabla_w f(x_s; w)\;\delta w \\
   &=& J_s \delta w,
\end{eqnarray}
where, $e_s$ is the reduction (growth) of error to (from) the learning equilibrium.  Taking the expectation over the training data set of size $S$, we obtain the average:
\begin{eqnarray}
   \epsilon &:=& <e_s>\\\nonumber
   &=&\frac{1}{S}\sum_{s=1}^S  e_s\\\nonumber 
   &=& <J_s> \delta w\\\nonumber
   &=:& {J}\delta w
\end{eqnarray}
Further, {\em wlog} assume a unit norm perturbation, $||\delta w||_2^2 = 1$ and define the metric ${G}:={JJ}^T$ (a symmetric positive semi-definite matrix) to obtain:
\begin{equation}
\epsilon^T {G}^{-1} \epsilon=1
\label{eq:metric}
\end{equation}

The interpretation of Equation \ref{eq:metric} is that it defines an elliptic locus of points (norm) of unit error length. The metric ${G}$ establishes preferred directions for error reduction. Akin to the {\em Manipulability}~\cite{yoshikawa85} metric in robotics or the {\em Visibility} metric in sensor planning~\cite{uppala2002viewpoint,ravela1994stealth}, which define the ease of motion in specific directions, $G$ defines the Learnability metric. It characterizes the efficacy of learning effort near-equilibrium along different dimensions. 

Solving an eigenvalue problem  ${G} U = U\Sigma^2$ and transforming $\nu = \Sigma^{-1}U^T\epsilon$ maps Equation \ref{eq:metric} to the equation of a circle: 
\begin{equation}
\nu^T\nu=1,
\label{eq:circle}
\end{equation}

Thus, a walk on the the unit circle $\nu(t)$, with implicit variable $t$, corresponds to a walk in two separate systems, ${G}_1$ and ${G}_2$ systems via the mappings $\epsilon_1(t) = U_1 \Sigma_1 \nu(t)$ and $\epsilon_2 = U_2\Sigma_2 \nu(t)$. 

The situation of interest is that the metric $G_1$ is a reference system using one set of parameters, and the metric $G_2$ is another system operated upon by an entirely different set of parameters, which may be unknown. For example, the first system might be (known) polynomial dynamics, and the second might be a neural network one wishes to parameterize. Clearly, if $G_1=G_2 \in \mathbb{R}^{n\times n}$ then $G_2$ behaves similarly in error space using an entirely different square-root, i.e., Jacobian map to its parameters. Unfortunately, it is not easy to directly recover the square-root $J_2\in \mathbb{R}^{n\times d_2}$ even if size  $d_2$ were known, which it is not. An alternate method is presented in the next section, which represents $G_2$ by size variables (and not explicit network parameters) and uses the balance between two systems to infer the size of the unknown system.

\section{Matched Spectral Features}
\label{sec:msf}
Our approach equates features (invariant scalar property) of Learnability metrics by using bounded training inputs, and expressing the Jacobians as a function of the number of system parameters and dimensions. This process is tractable, in contrast to inverting the metric for the parameter-Jacobian.  In particular, for the polynomial dynamics that are of interest,  consider the complete $(n,d)$ polynomial dynamics
\begin{equation}
\label{eq:fpoly}
    \dot{x}=f_{poly}(x;\alpha)\in \mathbb{R}^n
\end{equation}, with input variable $x=[x_1\ldots x_n]^T$. Expressing $f_{poly}$ as a sum of its $m = {n+d \choose d}$ monomials represented by parameter vector $\alpha=\left[\alpha_{1,0}\ldots \alpha_{1,m-1}, \ldots ,\alpha_{n,0}\ldots \alpha_{n,m-1}\right]^T$ of size $|\alpha|=n\;m$, one obtains: 
\begin{eqnarray}
    \dot{x}_i &=& \alpha_{i,0}+\sum_{k=1}^n \alpha_{i,k} x_k+ \sum_{l=n+1}^{m-1} \alpha_{i,l} z_{l},\; i=1\ldots n  \label{eq:polynet}
\end{eqnarray} The variable $z_{l}$ is simply a product of inputs of the form 
\begin{eqnarray}
   \label{eq:genpoly1}  z_{l} &=& \prod_{k=1}^n x_k^{d_{l}[k]},\\
   \label{eq:genpoly2} & s.t.\;\;&d_{l}\in [0,d]^{n\times 1} \\ \nonumber
     &&d_{l} \notin \left\{d_n\ldots d_{l-1}\right\}\\\nonumber
     &&1<\sum_{o=1}^n d_{l}[o] \leq d,
\end{eqnarray}
where, $d_n=\phi$. The fist two terms of Equation\;\ref{eq:polynet} cover the constant and first degree terms, the remainder generate all the monomials up to degree $d$. Either a  systematic sweep or random sampling generates  each unique  (see Equation~\ref{eq:genpoly2}) $n$-length vector $d_l$ in Equation\;\ref{eq:genpoly1} . All the ${n+d \choose d}-n-1$ $z_l$ terms  are generated once and reused in all system dimensions; if a particular monomial term ($i,l$) is absent, then $[\alpha_{i,l}]=0$.

This construct, together with sampling inputs from the $n$-dimensional bounded hypercube, that is the set $x_i\in \{-1, 1\}=:U, i=1\ldots n$, produces the trace:
\begin{eqnarray}
    tr({G}_{poly})&=&tr(\Sigma^2_{poly});\\
    &=&tr(J_{poly}J_{poly}^T);\\\nonumber
    &=&||J_{poly}||_{fr}^2;\\
    &= &n {n+d \choose d}.
\end{eqnarray}
Note that $tr(G_{poly})$ is fully determined by the size variables $(n,d)$, which we express as: $tr(G_{poly})$ reaches its size bound.

\paragraph{Equilibrium Learnability Trace [Frobenius Norm]:}
The trace of the Learnability metric $G$ is invariant to unitary transforms, and it is the Frobenius norm of the parameter-Jacobian $J$. It is also the system's ($f_{poly}$) local spectral property. In particular, the eigenvalues $[\sigma_{kk}^2]_{k=1}^n=:\Sigma^2$ are the square of the  singular values (sorted decreasing) of $J$. In contrast to an exact spectral match between two systems, the weak form prescribes only that the traces match. We call this feature the {\em  Equilibrium Learnability Trace} (ELT). It also has interpretations in terms of stability and Lyapunov exponent sums. 
\paragraph{Leading Lyapunov Exponent [Spectral Norm]:}
There are other possibilities. For example $\sigma^2_{11}$, the largest eigenvalue, corresponds to the local Lyapunov exponent a the learning dynamics. This is also the spectral norm of the Learnability metric. To see how, define the learning objective as a least squares problem\footnote{Extensions to other loss functions are possible but left out of the scope of this paper.} for a system near equilibrium, 
\begin{equation}
 y_s = f(x_s,w)+e_s,
\end{equation}
 as, 
\begin{equation}
    \mathcal{J}(w):= \frac{1}{2S}\sum_{s=1}^S (y_s - f(x_s; w))^2,
\end{equation}
and onsider the nominal gradient descent rule at iteration $j$:
\begin{eqnarray}
    w_{j} &=& w_j - \beta \frac{\partial \mathcal{J}}{\partial w_j},\;\;\beta>0,
\end{eqnarray}
starting from initial condition $w_0$, that upon convergence $w_M, M>0$ is an estimate of $w^*$. The gradient may be further expanded at iteration $j$ as
\begin{eqnarray}
        w_{j+1}&=& w_j + \frac{\beta}{S}  \sum_{s=1}^S  \nabla_{w_j} f(x_s;w_j)^T e_{js}.
\end{eqnarray}
Model $\nabla_{w_j} f(x_s;w_j) = \bar{J}_j +\tilde{J}_s$ with $\sum_s \tilde{F}_s = 0$. Further assume $e_{js} = \epsilon_j+\tilde{e}_s$ with $\sum_s \tilde{e}_s =0$, and assume the errors and parameters (gradients) are independent near equilibrium\footnote{This can be a poor assumption far away from equilibrium}, so that $\sum_s  \tilde{F}_s^T \tilde{e}_s= 0$. With these assumptions,the approximate update is: 
\begin{eqnarray}
    w_{j+1} &=& w_j + \beta  \bar{J}_j^T \epsilon_{j}.
    \label{eq:gradient}
\end{eqnarray}
Consequently, we may write the error growth/decay as
\begin{eqnarray}
    e_{j+1,s} &=& y_s-f(x_s;w_j + \beta  \bar{J}_j^T \epsilon_{j}),
\end{eqnarray}
which, to first-order expresses the discrete expected-error dynamics as 
\begin{eqnarray}
    \epsilon_{j+1} &=& (I_{nxn} -\alpha J_jJ_j^T )\epsilon_{j}= (I_{nxn} -\beta G_j )\epsilon_{j}\\\label{eq:learndyn}
    &\leq& (1-\beta\sigma^2_{11,j})\epsilon_{j}
\end{eqnarray}
Thus, $\beta\sigma^2_{11,j}> 1$ implies local stability. Note that Equation~\ref{eq:learndyn} follows from Equation~\ref{eq:metric} directly, but the present form shows explicit dependence between error reduction and weight updates. The leading eigenvalue $\sigma^2_{11}$ corresponds to the local leading Lyapunov exponent~\cite{Strogatz2015} of the continuous version of Equation~\ref{eq:learndyn}. Thus, over iterations, the integrated {\it Learnability metric's  Finite Time Lyapunov Exponent}~\cite{Haller2001DistinguishedFlows} (where time corresponds to iterations) is a feature (L-FTLE). Since $tr(G_j)$ is parameterized by size variables (the next section shows for neural networks,  previously described for polynomial system $f_{poly}$), so is $\sigma^2_{11,j}$. However, this paper uses ELT; the L-FTLE is left to a future paper. 

\subsection{Exact PolyNets}
 Matching  spectra allows sizing. Consider a one-hidden-layer PolyNet\cite{trautner2019} $f_\mathbf{PN}(x,w)$ ({\bf PN}) with $n$ inputs, $h_\mathbf{PN}\ge 1$ hidden nodes, and $n$ outputs trying to mimic $f_{poly}$ sized as $(n,d)$. Similar to two-layer network models of polynomials~\cite{Barron1993,Andoni2014}, the PolyNet contains multiplicative nodes~\cite{li02} and, like residual networks\cite{he16}, direct input-output connections. All input-hidden layer weights are $1$, and activation functions are set to identity, i.e. pass-through. The PolyNet has $nh_\mathbf{PN}+n^2+n$ active parameters and thus with inputs $x\in U^n$ hyper-cube $tr(G_\mathbf{PN}) = nh_\mathbf{PN}+n^2+n$. Balancing $tr(G_\mathbf{PN})=tr(G_{poly})$ yields:  
\begin{equation}
    h_\mathbf{PN}= {n+d \choose d} - n - 1.
\end{equation}
For example, with $n = 3, d=2$, $h_\mathbf{PN}=6$, which is the number of monomials of degree two. Equation\;\ref{eq:polynet} trivially defines the PolyNet; the first term represents output bias terms, the second term represents direct input-output connections, and the third term represents the $m-n-1$ hidden nodes without bias terms. The input-to hidden connections are all of weight $1$. A reading of the governing equation specifies the PolyNet. For this reason, $f_\mathbf{PN}$'s ELT ``reaches the bound" identically to the Frobenius norm of $f_{poly}$'s parameter-Jacobian $J_{poly}$. 

\begin{figure}[ht]
\centering
\includegraphics[width=0.9\linewidth]{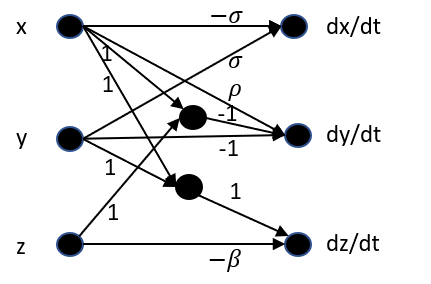}
\caption{A Continuous-time Neural Network with two hidden nodes and multiplicative units implements L63 exactly. }
\label{fig:lornet}
\end{figure}

\paragraph{Lorenz 63 Example}
We now apply the above procedure for a particular dynamical system. The Lorenz-63 ({ L63}) model~{\cite{Lorenz1963}} was originally used to describe 2-dimensional Rayleigh-B\'enard (RB) convection, in which the parameters of the streamfunction and temperature fields can be written in a set of ordinary differential equations {\cite{Lorenz1963}}: 
\begin{equation}
\begin{aligned}
    \dot{X} &= \sigma(Y - X);\\
    \dot{Y} &= \rho X - Y - XZ;\\
    \dot{Z} &= -\beta Z + XY, 
\end{aligned}
\label{eq:L63}
\end{equation}
where, $X$ and $Y$ are the strengths of the streamfunction and temperature modes, and $Z$ represents the deviation of the vertical temperature profile from linearity.Typical parameters are $\sigma = 10$, $\beta = 8/3$, and $\rho = 28$. The L63 system has only three free parameters, and its exact PolyNet ({\bf PNL63}) compiled from the equations requires only two hidden nodes, see Figure\;\ref{fig:lornet}. Given $h_\mathbf{PNL63}=2$, $n=3$, the equilibrium Learnability trace is $tr(G_\mathbf{PNL63})=6$, which is as obtained exactly from Equation\;\ref{eq:L63} because:  \begin{equation}
    J_{L63} = \begin{bmatrix} (Y-X)&0 &0\\
                                  0 & X & 0 \\
                                  0 & 0 & -Z\end{bmatrix}\stackrel{bound}{\rightarrow}\begin{bmatrix} \pm 2&0 &0\\
                                  0 & \pm 1 & 0 \\
                                  0 & 0 & \pm1\end{bmatrix},\\
\end{equation}
and therefore $J_{L63}$'s Frobenius norm $tr({G}_{L63})=6=tr(G_\mathbf{PNL63})$ matches {\bf PNL63}'s ELT. 

\paragraph{Discrete Neural Dynamics}
\begin{figure}[ht]
\centering
\includegraphics[width=0.95\linewidth]{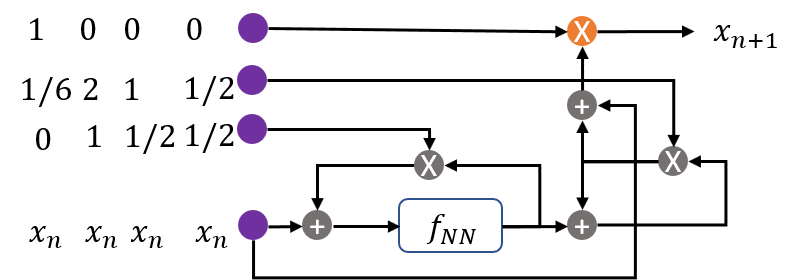}
\caption{A neural circuit for an explicit fourth-order Runge Kutta method for discrete-time solutions to continuous-time neural dynamics $f_{NN}$.}
\label{fig:rk4}
\end{figure}
Continuous-time ODEs (e.g.,\;Equation~\ref{eq:polynet})  are typically time-discretized using numerical time-stepping schemes. PolyNets constructed from the continuous ODEs are naturally amenable to numerical simulation due to the correspondence. Trautner et al.\cite{trautner2019} further show that a variety of time-stepping schemes such as Runge-Kutta\cite{BUTCHER20001} and Adams-Bashforth-Moulton\cite{BUTCHER20001} are themselves recurrent neural networks. Thus a recurrent neural circuit embedding the neural network corresponding to the continuous-time function implements the end-to-end discrete-time neural dynamical system. Switching between ``neural computation" and ``numerical simulation" is unnecessary. In the context of the present paper, the discrete neural dynamical system is also thus sized when the continuous-time version is. The parts that correspond to the time-stepping procedure are not parameters and ignored. 

\subsection{Classical One-Layer Neural Dynamics}
In contrast to the PolyNet $f_\mathbf{PN}$ matching $f_{poly}$, lets define a classical one layer network ({\bf CN}) as having $n$ input and output nodes each, $h$ hidden nodes, and only additive neurons with either semi-linear (e.g. ReLU) or saturating activations (e.g., tanh). The intent is to construct one similar to the two-layer ${\bf PN}$ network, minus the multiplicative neurons and pass-through activation.  Note that {\bf CN}'s ELT doesn't ``achieve the bound," in contrast to another network {\bf LN} that is identical except for containing pass-through activation (i.e., $h_\mathbf{CN}=h_\mathbf{LN})$. The {\bf CN} network contains attenuating and saturating nonlinearities. Thus,
\begin{equation}
\label{eq:bnd1}
    tr({G}_\mathbf{CN}) = c\;tr({G}_\mathbf{LN}),\;\;c\leq 1.
\end{equation} 
Nonlinear activation entails that the network parameters seep into  $J_\mathbf{CN}$ so that the $G_\mathbf{CN}$ becomes parameter-dependent and requires training to assess, but this is the situation we wish to avoid. Therefore, to equate 
\begin{equation}
 \label{eq:bnd2}
   tr({G}_\mathbf{CN}) = tr({G}_{poly}),
\end{equation} balance ${G}_\mathbf{LN}$ which reaches the bound with a scaled version of the $J_{poly}$'s Frobenius norm, i.e.,
\begin{equation}
 \label{eq:bnd3}
   tr({G}_\mathbf{LN}) = \frac{1}{c} tr({G}_{poly}),
\end{equation}
from which, one gets $h_\mathbf{CN}:=h_\mathbf{LN}$. This requires calibration of $c$. Without calibration of $c$ we obtain an inequality:
\begin{equation}
 \label{eq:bnd4}
  tr({G}_\mathbf{LN}) \geq  tr({G}_{poly}).
\end{equation}
Thus, $h_\mathbf{CN}$ is a lower bound. Permitting biases in the hidden and output nodes of {\bf CN}, we obtain: 
\begin{equation}
    h_\mathbf{CN} \geq \left\lceil \frac{n}{2n+1} \left[ {n+d \choose d}-1\right] \right\rceil.
\end{equation}

For the $n=3, d=2$ problem, we need $h_\mathbf{CN}\geq 4$ nodes. Fewer than four nodes would imply under performance, but an order of magnitude larger would imply over-fitting. 
\paragraph{Classical One-Layer with Residual Connections}
In addition to the connections in {\bf CN}, if $n^2$ skip connections ({\bf SC}) are permitted from input to output, the balance equation obtained is:
\begin{equation}
    h_\mathbf{SC} \geq \left\lceil\frac{n}{2n+1}\left[{n+d \choose d}-n-1\right]\right\rceil.
\end{equation}
For $n=3,d=2$, we obtain $h_\mathbf{SC}\geq 3$. In the next section, we shall see how to tighten the bound further. 

If one wished to design a single layer classical Feedforward Neural Network for L63 with only knowledge being $n=3,d=2$,i.e., the parameters are unknown, then the recommendation of at least four nodes ({\bf CN}) closely matches experimental results Li and Ravela\;\cite{Li2019bc} obtained. They design compact neural networks to emulate L63 and show that both systems have near-identical predictability beyond the lower bound. Remarkably, there are no particular assumptions about the data size or desired training error; the bulwark of other sizing approaches. Only bulk comparisons appear to provide a tighter bound than, for example, Andoni et al. provide.  The comparison, also reported in Li and Ravela\;\cite{Li2019bc} to learning from $n=3, d=2$ polynomial dynamics with a forward-Euler scheme with $h$ neurons and Root Mean Square (RMS) error target $\epsilon$ is bounded by $h = \Omega(n^{6d}/\epsilon^3)$ according to\;{\cite{Andoni2014}}. More than $5\times10^5$ nodes are needed when $\epsilon \sim 1$. Neither our results nor Li and Ravela's experiments support the result. 

\section{Random Network Spectra}
\label{sec:rns}

Ideally, both the upper and lower size bounds are desirable. However, being a function of the available data, the target error, and neural architecture, they remain difficult to quantify. Governing equations provide a considerable advantage in estimating the lower bound. However, addressing the general architectures' network-weight dependence is essential to tighten the ELT bound further. 

To address the issue, we adopt a Bayesian approach, quantifying the cumulative density $F_{d|h}(D\leq d | h)$ for  $n$-dimensional $d$-degree polynomial dynamics that a network of size $h$ can at best support. Inversion for the cumulative density $F_{h|d} (H_0<h | d)$  prescribes the size lower-bound as 
\begin{equation}
\label{eq:hlb}
    h^* = \argmax_h\;\lim_{p_0\rightarrow 1}\left\{F_{h|d} (H_0<h|d) < p_0\right\}. 
\end{equation} 

\begin{figure}[ht]
\centering
\includegraphics[width=1.\linewidth]{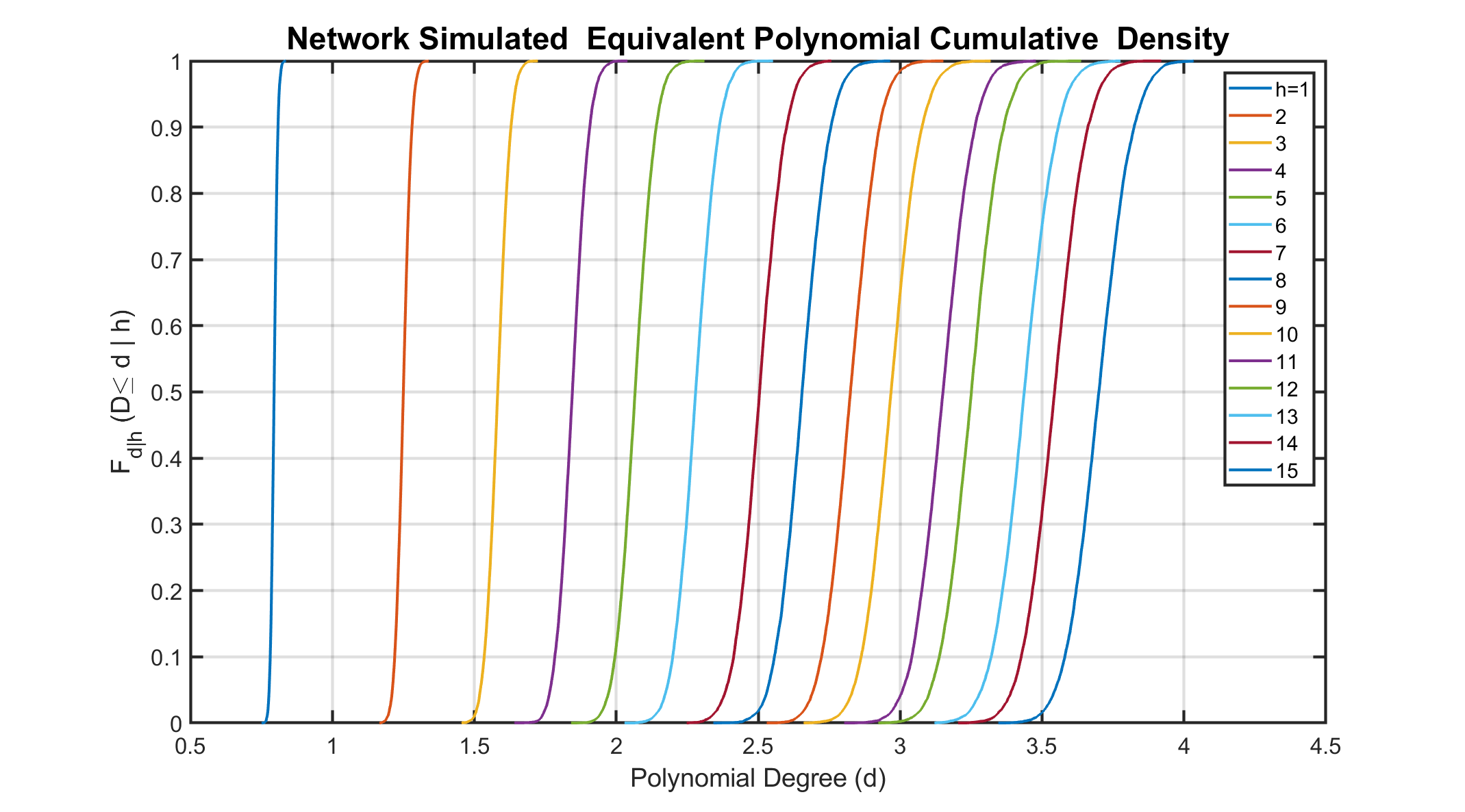}
\caption{Random one-layer networks with $tanh$ activation, bounded inputs and normally distributed weights yield cumulative density functions for the supported polynomial degree. }
\label{fig:netdegree}
\end{figure}
 To see how, consider the following steps described here for a {\bf CN} type network architecture. First, simulate random  {\bf CN} networks with bounded inputs from $U$, and random weights drawn i.i.d. from a Normal distribution\footnote{The weight choice is not restrictive; a single normalization is needed for reduction from other systems.}. Simulated  ELTs are fit to the generalized extreme value distribution~\cite{Haan2010} to estimate the empirical maximum (i.e., ELT with cumulative probability approaching $1$). Let's call this ELT $tr({G}^*_\mathbf{CN})$. The ELT $tr({G}_\mathbf{LN})$ of the corresponding {\bf LN} network, which is not weight dependent, determines the maximum supported $d^*$-degree  $n$-dimensional polynomial dynamics, and calibrates the constant $c$; see Equations~\ref{eq:bnd1}-\ref{eq:bnd4}. Thus, the equivalent polynomial degree $d$ for each particular simulated ELT is obtained, empirically quantifying cumulative density $F_{d|h}(D\leq d | h)$. Shown in Figure~\ref{fig:netdegree} are the cumulative densities for various network sizes for $n=3$ problem of a {\bf CN} type network with $\tanh (\cdot)$ activation.  Note that there is no learning involved in these steps. 
 
\begin{figure}[ht]
\centering
\includegraphics[width=1.\linewidth]{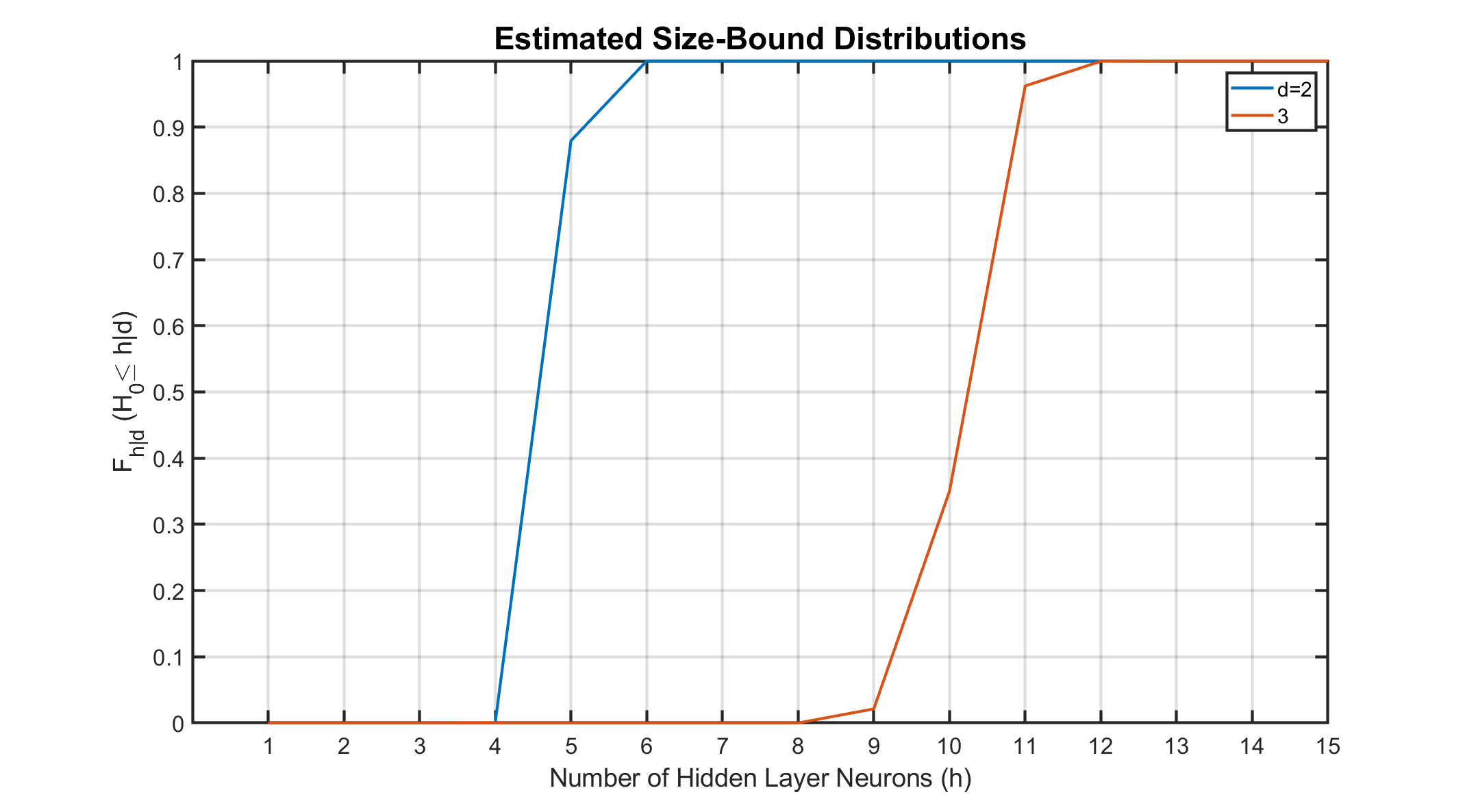}
\caption{Network Size Estimates of Effective Polynomial Degrees for $n=3$, and $d=2$ and $d=3$. At least six neurons are needed for $d=2$, and $13$ for $d=3$. }
\label{fig:netsize}
\end{figure}

 From $F_{d|h}(D\leq d | h)$, estimate the probability density $P_{d|h}=\frac{dF}{dh}$ and inverted it via Bayes rule to construct $P_{h|d}$, assuming uniform priors on $P(h)$ and $P(d)$ in our experiments.  This produces the conditional cumulative-density for the size lower-bound random variable $H_0$, i.e. $F_{h|D} (H_0<h | d)$, and Equation~\ref{eq:hlb} provides the necessary {\em lower-bound} value $h^*$. In Figure~\ref{fig:netsize} the cumulative densities are shown for $n=3$, $d=2$ and $d=3$. A numerical examination shows that $h^{(2)}_\mathbf{CN}$, the lower-bound for the new Bayesian estimated size variable for $n=3, d=2$ has risen as expected; $h^{(2)}_\mathbf{CN}\geq h^*_\mathbf{CN}=6$ for $n=3,d=2$. 
 
 Thus, modeling the {L63} system, if the perfect unknown parameters are known, $h_\mathbf{PNL63}=2$. The PolyNet for the full $n=3,d=2$ polynomial reveals $h_\mathbf{PN}=6$ nodes. The general network model {\bf CN} suggests $h_\mathbf{CN}\geq 4$ and the bound $h^{(2)}_\mathbf{CN}\geq 6$. Bayesian equivalent calculations for $h_\mathbf{SC}$ are similarly performed.  Although $h^{(2)}_\mathbf{CN}$ and $h_\mathbf{PN}$ are the same, the networks are quite different -- the latter is exact. 
 
 Note that even the Bayesian version calibrating $c$ prescribes a lower bound. That is because $h^*_\mathbf{CN}$ is an empirical estimate of the maximal ELT. It cannot be determined precisely. Thus, the bound is surprising from this perspective as well.  The lower bounds ($ 4, 6$) support  Li and Ravela's experimental results on lower-bounding the needed neurons to mimic polynomial dynamics with high fidelity. Our bounds confirm the absence of over-fitting.

\section{Conclusions}
\label{sec:concl}
This paper investigates sizing networks that seek to emulate polynomial dynamics and further adapt with data. Our work quantifies a Learnability route using the near-equilibrium behavior of learning systems in training. In the presence of governing equations (theory), it equates the near-equilibrium behavior of the learning and reference system under parameter perturbations. This Learnability metric quantifies this behavior and yields features that must match for comparable systems. The system dimensions and size variables bound the Frobenius norm of the Jacobian for bounded inputs and polynomial dynamics. Likewise, they also bound the Learnability metric for learning systems. Matching the two provides a method to size networks. Sizing is exact for PolyNets~\cite{trautner2019} and linearly activated networks. If the system dimension and degree lower bound the actual polynomial dynamics, these networks maintain the bound with no further loss. The linear networks also bound networks with nonlinear activations. Bayesian inference further tightens these bounds through simulation to calibrate the nonlinearities. One specific result of our work is that the size lower bound estimates of Lorenz system~\cite{Lorenz1963} confirm that Neural emulators of this system can learn to become chaotic with only a few neurons and data, without overfitting~\cite{Li2019bc}. 

\section*{Acknowledgments} 

The authors were members of the Earth Signals and Systems Group (ESSG), where this work was conducted. This paper is supported in part by the MIT UROP program, MIT Environmental Solutions Initiative, Liberty Mutual award 029024-00020, and  ONR  award N00014-19-1-2273. Its contents are solely the responsibility of the authors and do not necessarily represent the official views of the sponsors.
\bibliographystyle{plain}
\bibliography{secondary_general_project,newrefs}

\end{document}